 \documentclass[smallabstract,smallcaptions]{dccpaper}

\usepackage{cite}
\usepackage{graphicx}
\usepackage{url}
\usepackage{amsmath}
\usepackage{amssymb}
\usepackage[caption = false, font=footnotesize]{subfig}
\usepackage{float}
\usepackage{booktabs} 
\usepackage{multirow}
\usepackage[colorlinks=true,linkcolor=blue]{hyperref}%
\usepackage{array}
\usepackage{color}
\usepackage{tabularx}
\usepackage{longtable}
\usepackage{tabu}
\usepackage{pifont}
\usepackage{amssymb}
\usepackage{ragged2e}
\usepackage{tabularx}
\usepackage{longtable}
\usepackage{booktabs}
\usepackage{rotating}
\usepackage{makecell}
\begin{document}

\newcommand\blfootnote[1]{%
  \begingroup
  \renewcommand\thefootnote{}\footnote{#1}%
  \addtocounter{footnote}{-1}%
  \endgroup
}

\title
{
\textbf{Generative Face Video Coding Techniques and Standardization Efforts: A Review}
}

\author{%
Bolin Chen$^{\ast}$, Jie Chen$^{\dag}$, Shiqi Wang$^{\ast}$, and Yan Ye$^{\dag}$\\[0.5em]
{\small\begin{minipage}{\linewidth}\begin{center}
\begin{tabular}{ccc}
$^{\ast}$City University of Hong Kong & \hspace*{0.5in} & $^{\dag}$Alibaba Group \\
\end{tabular}
\end{center}\end{minipage}}
}

\maketitle
\begin{abstract}
Generative Face Video Coding (GFVC) techniques can exploit the compact representation of facial priors and the strong inference capability of deep generative models, achieving high-quality face video communication in ultra-low bandwidth scenarios. This paper conducts a comprehensive survey on the recent advances of the GFVC techniques and standardization efforts, which could be applicable to ultra low bitrate communication, user-specified animation/filtering and metaverse-related functionalities. In particular, we generalize GFVC systems within one coding framework and summarize different GFVC algorithms with their corresponding visual representations. Moreover, we review the GFVC standardization activities that are specified with supplemental enhancement information messages. Finally, we discuss fundamental challenges and broad applications on GFVC techniques and their standardization potentials, as well as envision their future trends. The project page can be found at \url{https://github.com/Berlin0610/Awesome-Generative-Face-Video-Coding}.

\end{abstract}

\vspace{-1.2em}
\section{Introduction}
\vspace{-0.6em}
In recent years, face video communication has shown explosive growth and great application potentials, which can support the service needs of online video conferencing/chat, education, e-commerce, live broadcasting and other industries. In particular, visual face data in these multimedia applications has become an important type of data occupying large proportion of bandwidth in the transmission networks around the world. As such, the central problem is how to compactly characterize and efficiently transmit the face visual information for high-fidelity and high-quality communication.

In the past decades, a wide variety of video codecs, such as H.264/Advanced Video Coding (AVC)~\cite{wiegand2003overview}, H.265/High Efficiency Video Coding (HEVC)~\cite{sullivan2012overview} and H.266/
Versatile Video Coding (VVC)~\cite{bross2021overview}, have been developed and optimized towards the trade-offs between reconstruction quality and bit-rate constraints. Although these video codecs have enabled the applications of video conferencing/chat and live broadcasting, there is still large room for improvement since these hybrid coding frameworks are not designed with the particular statistical characteristics of face visual signal. In addition, Model-Based Coding (MBC)~\cite{7268565,1457470,305867,lopez1995head,150969,364463,1230217}, which could date back at least to the 1950's, was dedicated to boosting the face video compression efficiency via strong face priors, but was encumbered by the poor-quality face synthesis until the recent rise of deep learning technologies. Inspired by the recent progress of deep generative models~\cite{VAE,goodfellow2014generative,NEURIPS2021_49ad23d1}, especially for Generative Adversarial Networks (GAN), the poor-quality face reconstruction of early MBC technologies can now be well remedied and improved. In particular, learning-based face reenactment/animation models~\cite{FOMM,siarohin2021motion,hong2022depth} have fulfilled the great promises of generative face video compression (GFVC)~\cite{facebook2021,ultralow,9859867,CHEN2022DCC,icip2022zhao,chen2023csvt,chen2023interactive}, where the analysis-synthesis model can economically represent motion parameters at the encoder side and reconstruct high-quality face video at the decoder side. As such, ultra-low bitrate visual face video communication can be realized. 

GFVC has shown great potentials in efficient coding performance and intelligent applications that are difficult to achieve for traditional codecs. In this paper, we review the recent technical progress and standardization activities regarding generative face video compression, targeting at ultra-low rate communication scenarios, user-specified animation/filtering and metaverse-related functionalities. In particular, we generalize a series of compression methodologies with different visual representations into a general GFVC framework. Furthermore, we introduce the technical evolution from the early MBC to the latest GFVC with the Artificial Intelligence (AI) powered technologies and the related standardization efforts. Finally, we envision promising applications and discuss existing challenges when utilizing this state-of-the-art compression technique. 

\begin{figure*}[tb]
\centering
\vspace{-3.8em}
\subfloat[General Framework]{\includegraphics[width=0.58\textwidth,height=3.5cm]{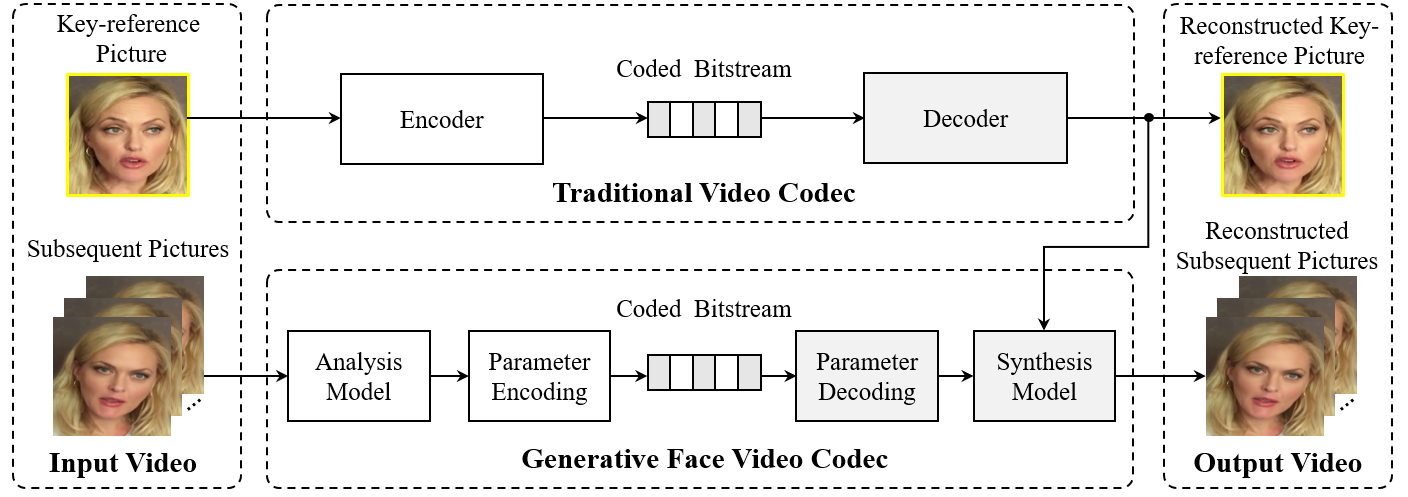}}
\hspace{1.2em}
\subfloat[Facial Representations]{\includegraphics[width=0.36\textwidth,height=3.4cm]{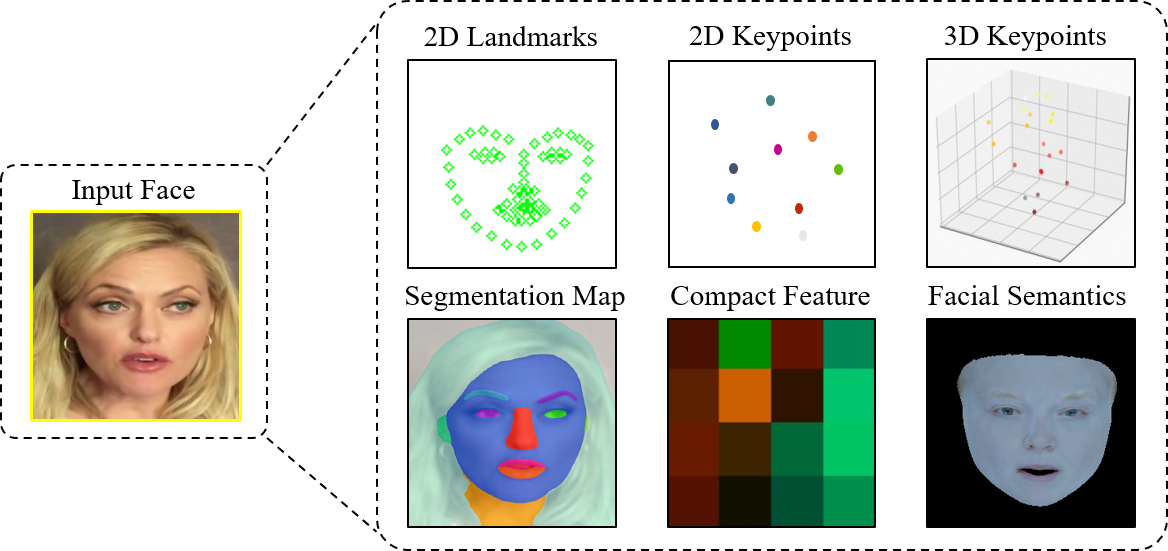}}
\vspace{-0.6em}
\caption{Illustration of the general framework and different facial representations for the GFVC system.} 
\vspace{-1.2em}
\label{fig1} 
\end{figure*}

\vspace{-1.5em}
\section{Generative Face Video Coding}
\vspace{-0.43em}

Herein, we give a progress overview of face video compression techniques. In particular, both the early MBC and the latest GFVC are introduced, including the technical details and compression performance.

\begin{table*}[p]
\vspace{-1.5em}
\renewcommand\arraystretch{1.35}
\caption{Summary of GFVC methods with different facial representations}  
\label{table1}
\centering
\resizebox{1\textwidth}{!}{
\begin{tabular}{lllm{8.5cm}}
\hline
\textbf{Method} & \textbf{Published Time} & \textbf{Facial Representation} & \textbf{Distinguished Characteristics}\\ \hline
FOMM~\cite{FOMM} & NeurIPS 2019  & 2D Keypoints & Learns complex facial motion with 2D keypoints and its affine translation expressions without using any  prior information, which is the basis of keypoint-based GFVC algorithms.  \\
DAC~\cite{ultralow}          & ICASSP 2021  & 2D Keypoints     & Designs a FOMM-based video conferencing codec and achieves the promising RD performance.\\
VSBNet~\cite{9455985}        & ICMEW 2021   & 2D Landmarks     & Develops the Visual-Sensitivity-Based network with auxiliary frames and confidence score map to improve face fidelity.    \\
Face\_vid2vid~\cite{wang2021Nvidia} & CVPR 2021    & 3D Keypoints     & Proposes a neural talking-head video synthesis model that can achieve free-view control and interaction of face video without the aid of 3D graphics model, and can be further applied to video conferencing at ultra-low bitrate. \\
Mob M-SPADE~\cite{oquab2021low}   & CVPRW 2021   & Segmentation Map & Conducts comparative analyses of leading GFVC approaches and ultilizes the SPADE architecture and segmentation maps to develop the first real-time GFVC system on the mobile CPU.                            \\
SNRVC~\cite{9810784}         & DCC 2022     & Facial Semantics & Disentangles face motion information with a series of specific semantics for reconstructing face video under the low-bitrate constraints.          \\
CFTE~\cite{CHEN2022DCC}          & DCC 2022     & Compact Feature  & Transforms the temporal evolution of face video into compact feature representation based upon an end-to-end inference framework.  \\
C3DFD~\cite{li9810765}         & DCC 2022     & Facial Semantics & Designs an ultra-low bitrate digital human character communication paradigm via compact 3D face descriptors.    \\
MAX-RS~\cite{volokitin2022neural}        & CVPRW 2022   & 2D Keypoints     & Exploits multiple source frames to explore multi-view neural face video compression with view aggregation and selection.     \\
DMRGP~\cite{icip2022zhao}         & ICIP 2022    & Compact Feature  & Proposes a dynamic multi-reference prediction mechanism for GFVC to achieve the conversion from large head motion to small motion.       \\
HDAC~\cite{konuko2022hybrid}          & ICIP 2022    & 2D Keypoints     & Develops the layered coding scheme to improve long-term dependencies and alleviate background occlusions.       \\
CVC\_STR~\cite{2022bmvc}      & BMVC 2022    & 2D Keypoints     & Designs a frame interpolator to reduce the transmitted bitrate and a patch-wise super-resolution algorithm to improve the reconstruction quality.                                                                 \\
Bi-Net~\cite{9859867}        & ICME 2022    & 2D Keypoints     & Proposes a bitrate-adjustable hybrid GFVC framework with pixelwise bi-prediction, low-bitrate-FOMM and lossless keypoints.    \\
CTTR~\cite{chen2023csvt}          & TCSVT 2023   & Compact Feature  & Improves the learning ability to characterize the dynamic trajectories of face signal for better temporal consistency via the spatial-temporal adversarial training. \\ 
IFVC~\cite{chen2023interactive}          & arXiv 2023   & Facial Semantics & Develops the internal-dimension-increase-based mechanism to allow humans' interaction with the intrinsic visual representations, such as head translation, rotation, eye blinking and mouth motion.          \\
RDAC~\cite{konuko2023predictive}    & ICIP2023   & 2D Keypoints + Residual Map & Proposes a predictive coding scheme that can compress the residual information for quality improvement. \\\hline
\end{tabular}
}
\end{table*}
\vspace{-0.6em}
\subsection{Research Progress}
\vspace{-0.43em}
As for the visual compression of face-centric videos, early MBC techniques ~\cite{7268565,150969,364463,1457470,285624,lopez1995head,305867,201932} provided reasonable solutions regarding how to facilitate efficient image/video coding via their statistical priors. More specifically, the structural information in face images can be economically exploited as transmitted symbols (\textit{e.g.,} semantic parameters or facial edges) with the aid of the analysis model, and subsequently the decoded symbols can be utilized to reconstruct face images via the synthesis model. For example, Hotter~\cite{285624} designed an object-oriented analysis-synthesis codec that could compactly characterize the motion, shape and color of the moving object via the image analysis technique. Clark~\textit{et al.}~\cite{201932} proposed to capture three-dimension motion and shape from the object via the three-dimension face model. Also, Chowdhury \textit{et al.}~\cite{305867} adopted generalized cylindrical models and a 3D matching technique to characterize the motion information for moving objects. In addition, Lopez \textit{et al.}~\cite{lopez1995head} utilized the 3D head model to achieve face-specific video compression by encoding the relevant 3D facial parameters. Moreover, some mature techniques like Principal Component Analysis (PCA) and 3-Dimensional Morphable Model (3DMM) were applied into this task. However, limited by the inadequate synthesis technique at that time, the overall reconstruction quality was poor, thus resulting in this MBC direction not being widely developed. 

Recent progress in deep generative models, such as Variational Auto-Encoding (VAE)~\cite{VAE}, GAN~\cite{goodfellow2014generative}, Diffusion Model (DM)~\cite{NEURIPS2021_49ad23d1} have shown great potentials in high-quality synthesis, which can alleviate the unsatisfactory quality of early MBC techniques. In particular, face animation/reenactment models ~\cite{FOMM, siarohin2021motion,hong2022depth} are capable of economically characterizing the input frames with compact facial representations (\textit{e.g.,} 2D landmarks, 2D keypoints, 3D keypoints, compact feature, segmentation map and facial semantics as shown in Figure~\ref{fig1} (b)) and leverage the powerful learning capabilities of deep generative models to reconstruct these face frames, which have greatly advanced the progress of generative face video compression. Figure~\ref{fig1} (a) illustrates the general encoding-decoding processes of GFVC algorithms. At the encoder side, the key-reference frames of face video are compressed by traditional coding technique such as HEVC and VVC, and the subsequent inter frames are characterized with the compact transmitted symbols and coded into the bitstream. At the decoder side, the decoded key-reference frame and compact facial information are jointly fed into the synthesis model for reconstructing the video. In this manner, face video communication can be actualized towards ultra-low bitrate and high-quality reconstruction. In Table~\ref{table1}, we further summarize all representative work in this field and categorize these GFVC algorithms with different facial representations. To conclude, these GFVC algorithms have been developed towards more compact facial representation for reduced coding bits, more realistic image reconstruction for better perceptual quality and interpretable bitstream editing for wider application scenarios.

\begin{figure}[tb]
\centering
\vspace{-4.4em}
\subfloat[Rate-DISTS Performance]{\includegraphics[width=0.43\textwidth,height=3.6cm]{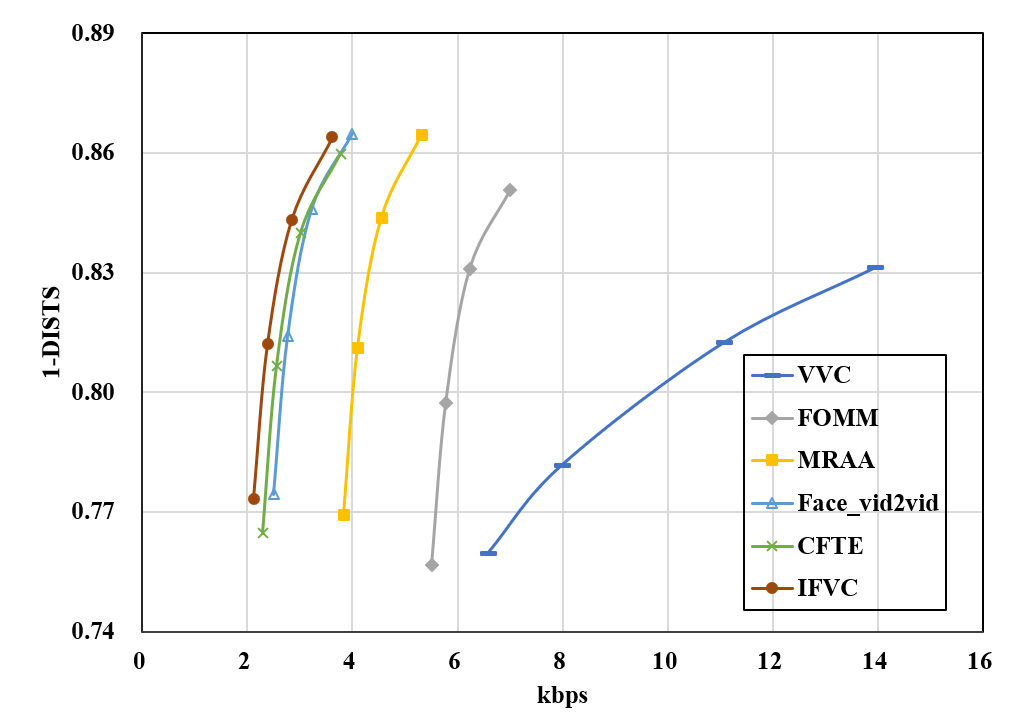}}
\subfloat[Rate-LPIPS Performance]{\includegraphics[width=0.43\textwidth,height=3.6cm]{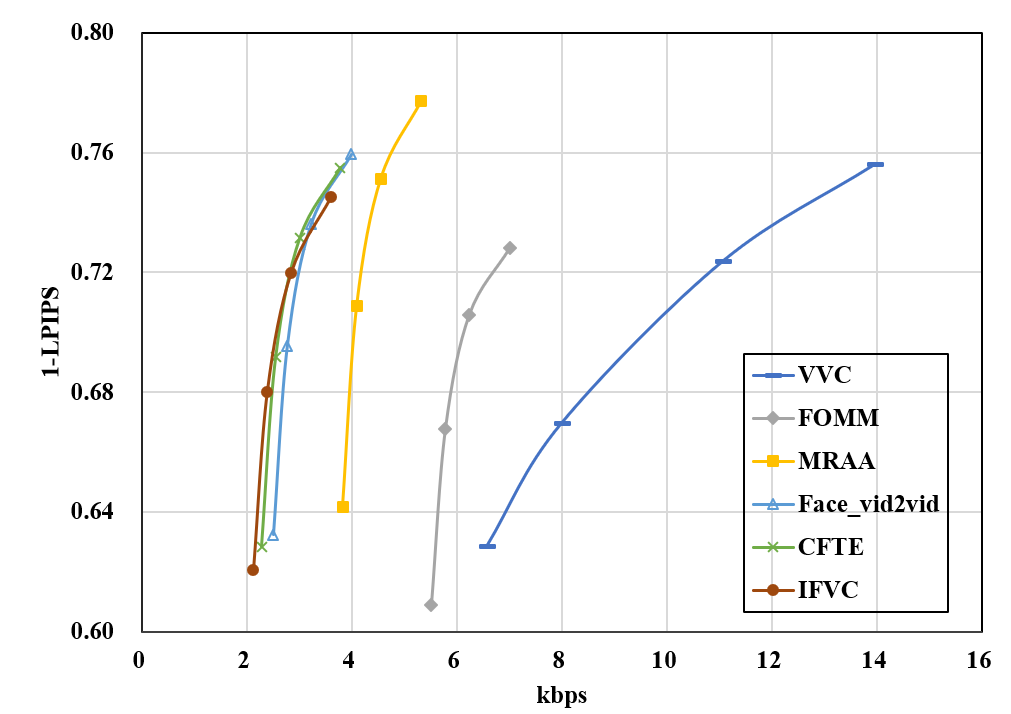}}
\vspace{-0.6em}
\caption{Rate-distortion performance comparison with the traditional codec (VVC~\cite{bross2021overview}) and generative compression systems (FOMM~\cite{FOMM}, MRAA~\cite{siarohin2021motion}, Face\_vid2vid~\cite{wang2021Nvidia}, CFTE~\cite{CHEN2022DCC} and IFVC~\cite{chen2023interactive}). } %
\vspace{-1.2em}
\label{fig4} 
\end{figure}

\vspace{-1.0em}
\subsection{Overall Performance}
\vspace{-0.55em}
To validate the overall rate-distortion performance of these above GFVC methodologies, we carry out the related experiments among the traditional codec and generative compression systems via the common test conditions and dataset described in ~\cite{chen2023interactive}.
\vspace{-0.5em}
\begin{itemize}
\item{For the testing dataset, there are 50 talking face video sequences selected as testing sequences from the VoxCeleb dataset ~\cite{Nagrani17}. Each of these testing sequences is cropped from the original sequence into 10-second long (i.e. 250 frames at 25 fps) with the resolution of 256$\times$256. The coded color format is RGB 444.}
\vspace{-0.8em}
\item{For common test conditions, the Low-Delay-Bidirectional (LDB) configuration in VTM reference software is adopted to encode these testing sequences, where the quantization parameters (QP) are set to 45, 47, 50 and 52. In addition, for generative face video compression scheme, its key-reference picture is compressed as an I picture using the VTM 10.0 reference software~\cite{JVET-S2002}, and other pictures are further represented with a series of facial parameters and compressed via a context-adaptive arithmetic coder. In particular, the QPs of the key-reference picture are set to 37, 42, 47 and 52.}  
\vspace{-0.6em}
\end{itemize}

Figure~\ref{fig4} provides the overall rate-distortion performance of the 50 test sequences for VVC, FOMM, MRAA, Face\_vid2vide, CFTE and IFVC. Since GFVC methods do not optimize for pixel-level distortion fidelity, traditional measures like Peak Signal-to-Noise Ratio (PSNR) and Structural Similarity Index (SSIM) are not suitable to measure the quality of the reconstructed video. Therefore, two well-known learning based quality metrics, Deep Image Structure and Texture Similarity (DISTS)~\cite{dists} and Learned Perceptual Image Patch Similarity (LPIPS)~\cite{lpips}, are used instead. Figure~\ref{fig4} illustrates that all generative compression methods can significantly outperform the latest VVC codec, and they can operate at ultra-low bit rate range that are difficult to reach for traditional codecs. %

\vspace{-1.2em}
\section{Standardization Efforts}
\vspace{-0.6em}
From 1990's, the ISO/IEC Moving Picture Experts Group (MPEG) began to push forward the standardization of the MBC technique in the MPEG-4 part 2 standard~\cite{MPEG4} since the H.261 standard couldn't provide satisfactory picture quality at the very low bit-rates ranges required for mobile transmission at the time. Compared with the block-oriented H.261 scheme, the MBC in MPEG-4 standard could better adapt to the local image characteristics and object motion compensation. Specifically, MPEG-4 proposed to separately encode each object and then multiplex them into a single bitstream at the encoder side, while the bitstream could be further de-multiplexed and decoded for the final image fusion at the decoder side. In addition, researchers~\cite{836279} also investigated the combination of the MBC and hybrid block-based schemes, where the model-based coder was incorporated into the block-based video coder for object prediction and coding. The corresponding motion parameters can be represented by facial animation parameters via the MPEG-4 standard. Moreover, MPEG-7~\cite{MPEG7} is another standard that proposes content description representation to allow human-computer interaction for multimedia content. However, limited by the capabilities of earlier image analysis-synthesis techniques, the standardized MBC technique in MPEG-4 part 2 has not even seen wide deployment for a long time.

Recently, inspired by Artificial Intelligence Generated Content (AIGC), a series of technical proposals on generative face video compression have been submitted to the Joint Video Experts Team (JVET) of the ISO/IEC SC 29 and ITU-T SG16. These proposals have mainly focused on the format of Supplemental Enhancement Information (SEI) messages. As the name indicates, SEI messages are additional and supplemental data that may be inserted into a coded video bitstream to enhance the use of transported video for a wide variety of purposes. The unique design of GFVC that employs the hybrid codec and extracts a series of facial parameters provides great possibilities to be standardized as SEI messages that accompany a video coding specification. Herein, a series of contributions have been made to the JVET group to advocate adding the corresponding syntax into VVC bitstreams in order to support generative face video coding.

\begin{figure*}[tb]
\centering
\vspace{-4.5em}
\subfloat[GFV SEI for Face Generation]{\includegraphics[width=0.42\textwidth,height=4cm]{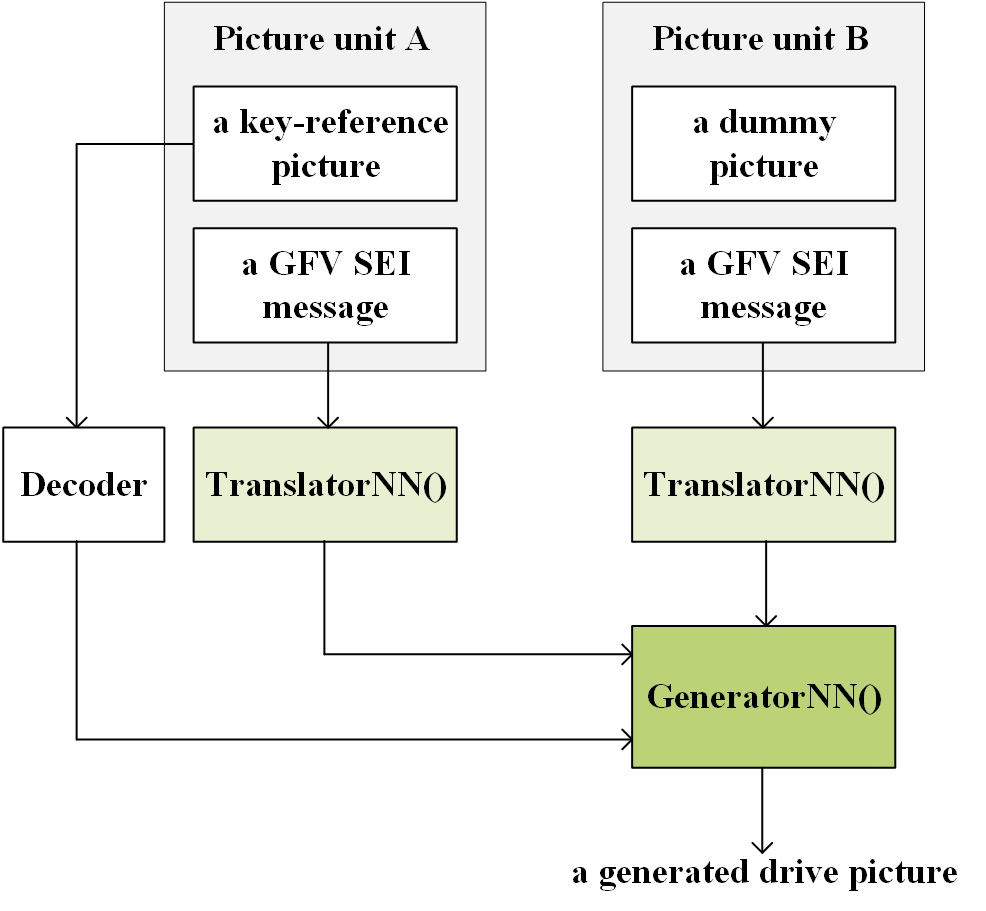}}
\hspace{0.3em}
\subfloat[GFV SEI for Face Fusion]{\includegraphics[width=0.48\textwidth,height=4cm]{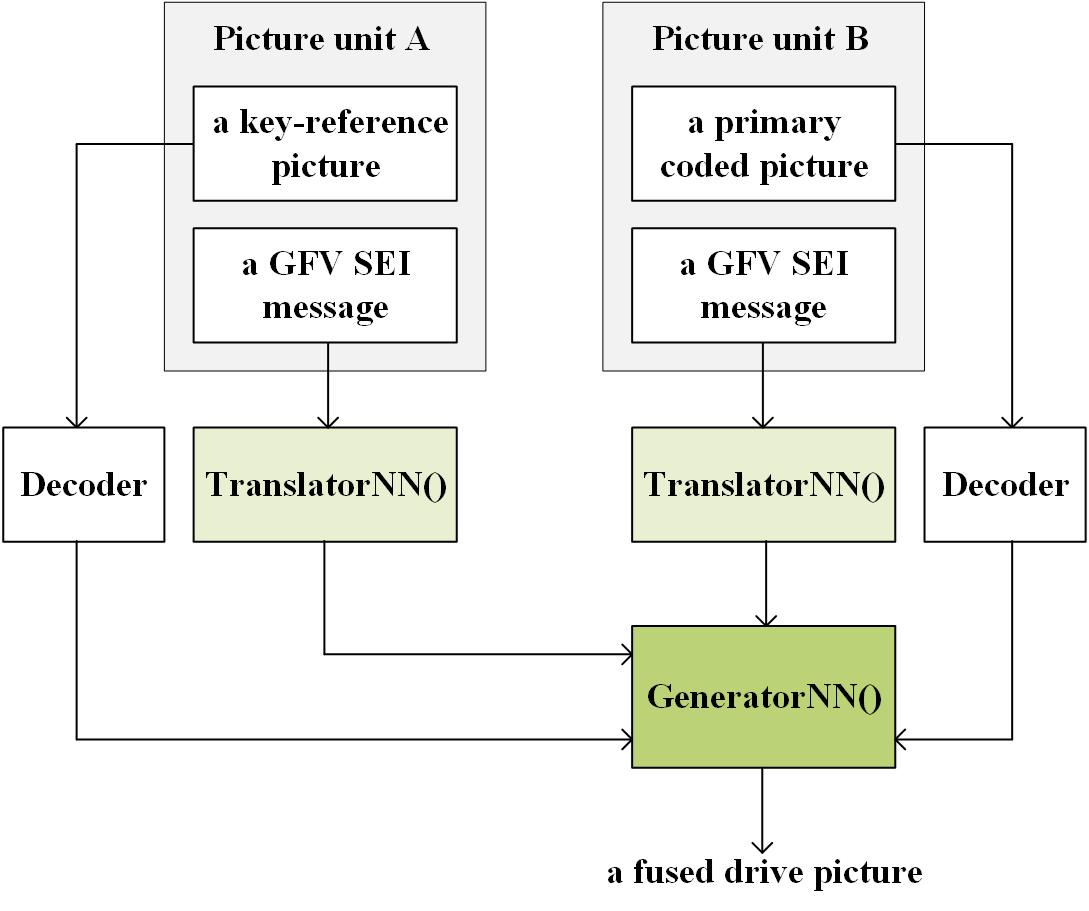}}
\vspace{-0.6em}
\caption{The processing order of GFV SEI defined in JVET-AF0234~\cite{JVET-AF0234}, where the parameter translator and the synthesis model are denoted as TranslatorNN() and GeneratorNN() respectively. It should be noted that TranslatorNN() can be bypassed when encoder (analysis model) and decoder (synthesis model) are matched.} 
\vspace{-1.2em}
\label{sei} 
\end{figure*}

More specifically, at the 29th JVET meeting, JVET-AC0088~\cite{JVET-AC0088} first proposed a generative face video (GFV) SEI message to allow VVC-coded pictures to be used as base pictures and add a small amount of bit overhead to represent facial semantics, which could enable ultra-low bit rate face video communication. Furthermore, the proposal JVET-AD0051~\cite{JVET-AD0051} to the 30th JVET meeting generalized the GFV SEI message to include syntax elements that correspond to a variety of facial representations, such as 2D keypoints, 2D landmarks, 3D keypoints, facial semantics, compact temporal feature and other formats. At the 31st JVET meeting, JVET-AE0080~\cite{JVET-AE0080}, JVET-AE0083~\cite{JVET-AE0083} and JVET-AE0280~\cite{JVET-AE0280} proposed to support more common syntax design of GFV SEI message and define the interface between the decoder (which decodes the key-reference pictures and parses the SEI message) and the generative neural network. In particular, JVET-AE0083~\cite{JVET-AE0083} also provided clear information on available open-source generative models as well as an example software implementation that could link the VTM software with a generative network. At the 32nd JVET meeting, the proposal JVET-AF0146~\cite{JVET-AF0146} sought to introduce refinements and grouping of 3D facial landmarks syntax for the GFV SEI message from JVET-AE0280. In addition, the proposal JVET-AF0048~\cite{JVET-AF0048} attempted to use a flow translator or a parameter translator to improve the interoperability between encoder and decoder that are not trained in an end-to-end manner. Experimental results showed that although reconstruction quality is somewhat lower in the case of “mismatched” encoder and decoder compared to matched (i.e., end-to-end trained) encoder and decoder, the quality is still acceptable if a flow translator or a parameter translator is used. Moreover, the proposal JVET-AF0234~\cite{JVET-AF0234} updated the common text of GFV SEI message proposed in JVET-AE0280 by introducing the parameter translator proposed in JVET-AF0048. The processing order of GFV SEI message with a parameter translator is illustrated in Figure~\ref{sei}.  

To conclude, the current effort to standardize GFVC techniques has been focused on the design of SEI messages and with the development goals towards rich face representations as well as effective decoder interoperability. Moreover, based on the suggestions from Ye \textit{et al.}~\cite{m64987} and discussions in the JVET group, a new Ad hoc Group has been established for comprehensive GFVC investigations from the perspective of software implementation, coordinating experimentation, interoperability study and so on.

\vspace{-2.5em}
\section{Applications and Challenges}
\vspace{-0.6em}
These GFVC techniques relying on a powerful deep generative network can enable the following application cases currently unavailable in the VVC standard,
\vspace{-0.5em}
\begin{itemize}
\item{ \textbf{Ultra low rate communication}: The key-reference pictures can be sent with large time gap in between, and in between two consecutive key-reference pictures, only a few facial parameters per time instance need to be sent, significantly reducing the coded bit rate compared to coding all pictures using VVC. This could warrant the service of more face video communication scenarios, including video conferencing/telephone, online broadcasting and live entertainment, where high-quality communication can be achieved even under the poor network bandwidth conditions.}
\vspace{-0.8em}
\item{ \textbf{User-specified animation/filtering}: Face animation and facial filters have become very popular in many mobile apps. As the encoder has the uncoded face images, it can extract more precise facial landmarks/keypoints and send them to the decoder together with the coded video bitstream, resulting in higher quality facial animation at the decoder side as shown in \href{https://github.com/Berlin0610/Awesome-Generative-Face-Video-Coding}{this project page}. This could allow the sender to have better control over the quality of animated or filtered facial pictures.} 
\vspace{-0.8em}
\item{ \textbf{Metaverse-related functionalities}: The emerging metaverse needs high-efficiency visual data compression techniques to support versatile communication. In particular, the existing GFVC techniques can provide the face composition in 3D space, such as the facial mesh reconstruction and 3D facial motion editing. Therefore, human beings can be more naturally allowed to interact with the virtual world, resulting in augmented immersive experience with reality.} 
\end{itemize}

Although the promising rate-distortion performance can be achieved by these GFVC approaches and the related face video communication scenarios can be better actualized along these veins, there exist drawbacks and challenges for their standardization as follows,
\vspace{-0.5em}
\begin{itemize}
\item{ \textbf{Unstable reconstruction quality}: Limited by the ability of deep generative models and the internal GFVC mechanisms (e.g., texture provided by the key-frame reference and compact representation for motion and posture), the reconstruction face pictures sometimes may contain objectionable distortions in background region, and/or inaccurate local motions in the mouth and/or eyes. Such instability in reconstruction quality may damage the feasibility of practical applications to a certain extent. Therefore, customized encoding techniques specifically designed for GFVC are needed, e.g. sending more the key-reference pictures more frequently when larger motion is present in the video, to alleviate issues with objectionable reconstructed quality.} 
\vspace{-0.8em}
\item{ \textbf{High decoder complexity}: The current model complexity for the GFVC decoder is much higher than that of a VVC decoder (or any other decoder based on traditional video coding framework). However, with careful simplifications to the generative networks, it had been shown in~\cite{oquab2021low} that real-time decoding on mobile platforms is already achievable. With more dedicated hardware acceleration targeted toward neural networks being deployed on mobile platforms, it is anticipated that more power-efficient and computing-efficient can be achieved in the near future. }
\vspace{-0.8em}
\item{ \textbf{Lack of appropriate perceptual evaluation}: The common quality metrics in video coding, such as PSNR and SSIM are not suitable to measure the GAN-based reconstruction (or any other reconstruction based on generative methods). Although DISTS and LPIPS currently used in the GFVC task are proposed in the feature domain, they lack consideration of strong facial priors. As a result, appropriate perceptual evaluation measures like~\cite{li2023perceptual} are highly required to more accurately evaluate the efficacy of the GFVC task. }
\vspace{-0.8em}
\item{ \textbf{Inadequate model interpretability}: The current GFVC techniques highly rely on the strong inference capability of deep generative models. Like other work based on deep neural networks, one key issue is to better understand the inner working of the model, e.g., why and how the inner mechanism and decision-making process of the generative model can well support the GFVC task. Gaining such understanding will allow fine tuning of the generative models and the encoding methods to further improve the GFVC performance in the same way that has been done to the traditional codecs for the past decades.}
\end{itemize}

\vspace{-1.8em}
\section{Conclusion}
\vspace{-0.8em}
In this paper, we have provided a thorough review of the recent progress and standardization efforts on generative face video coding. By reviewing the technical and standardized development from the early MBC in 1950's to the latest GFVC, it is clearly shown that compact prior representation and strong image synthesis ability both play vital roles in achieving promising rate-distortion performance. This advanced technique has shown great potentials in a wide variety of face video applications, whilst new applications and requirements in the post-COVID-19 era and the emerging metaverse realm also drive the rapid development of the generative face compression technique.

\vspace{-1.8em}
\section*{References}
\vspace{-0.5em}
\bibliographystyle{IEEEtran}
\bibliography{reference}
\vspace{-0.8em}
\end{document}